# Probabilistic index maps for modeling natural signals

**Nebojsa Jojic**
Microsoft Research
Redmond, WA

**Yaron Caspi**
Hebrew University

**Manuel Reyes-Gomez**
Columbia University
New York

## Abstract

One of the major problems in modeling natural signals is that signals with very similar structure may locally have completely different measurements, e.g., images taken under different illumination conditions, or the speech signal captured in different environments. While there have been many successful attempts to address these problems in application-specific settings, we believe that underlying a large set of problems in signal representation is a representational deficiency of intensity-derived local measurements that are the basis of most efficient models. We argue that interesting structure in signals is better captured when the signal is defined as a matrix whose entries are discrete indices to a separate palette of possible measurements. In order to model the variability in signal structure, we define a signal class not by a single index map, but by a probability distribution over the index maps, which can be estimated from the data, and which we call probabilistic index maps. The existing algorithms can be adapted to work with this representation. Furthermore, the probabilistic index map representation leads to algorithms with computational costs proportional to either the size of the palette or the log of the size of the palette, making the cost of significantly increased invariance to non-structural changes quite bearable. We illustrate the benefits of the probabilistic index map representation in several applications in computer vision and speech processing.

## 1  Introduction

In previous work, a very interesting step in the direction of color-invariance was made by Stauffer et al, who replace the image intensities with a self-similarity measure [5, 6]. They build a large "co-occurrence matrix" with an entry for every pair of pixels. This statistic is computed from a labeled training set, and as far as we know their technique is only used in supervised algorithms. The major problem is the size of the matrix ($10^5 \times 10^5$ entries for a $256 \times 256$ image), leading to computational and storage problems that have so far limited their experiments to tasks that use small images, e.g., pedestrian detection. Our representation is considerably more efficient and is easily used in unsupervised algorithms. It is also easily combined with other causes of variability in graphical models, e.g., the models developed by Jojic and Frey [2, 3]. However, as our experiments show, our new representation provides much greater color- and feature-invariance, which helped it outperform the appearance-based models in unsupervised transformation-invariant clustering tasks.

In speech applications, the most similar approaches, at least on the surface, seem to be mixture tying examples. However, these are typically used to reduce the amount of training examples needed for learning, and do not in fact provide invariance to local measurements as in the example we give in this paper. Our model assumes that the palette of measurements in a particular frequency band can change from utterance to utterance of the same word, but the indices into the palette are drawn from a single distribution defined by an HMM model.

## 2  Palette indexing

One efficient representation of an image is as the collection of indices, one index per pixel, that points to a separate table of possible values the pixels can take. This representation is heavily used in image formats, as it drastically reduces the storage requirements. Although the goal is usually storage efficiency, this is achieved by exploring self-similarity in the image, at least at the lowest level, which in itself is interesting for other image processing tasks, including image understanding. The index table is typically a color table, or a palette, but there is no reason why it would not contain derived features, such as wavelet coefficients, for example, in which case the representation may have a higher value in image processing tasks. The palette can be shared across a collection of images, e.g. a video sequence, or a set of objects with similar structure, but different colors.

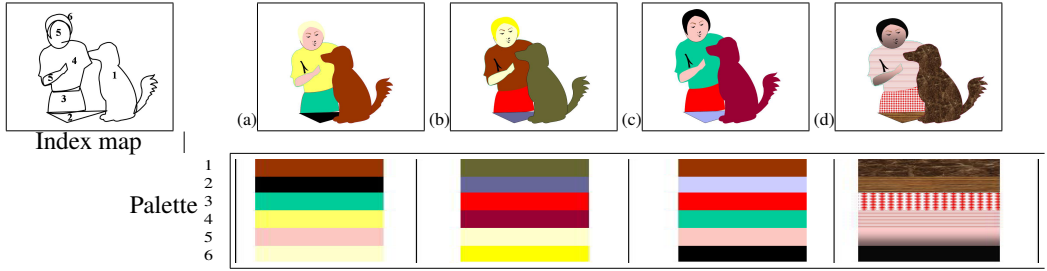

Figure 1: Illustration

Furthermore, the same strategy for splitting the variability in the data between the palette of possible measurements and the index table associated with the data is applicable to other natural signals, such as audio, for example.

In this section, we discuss possible assumptions on dependencies among variables in such representations to point out the similarities and differences between standard techniques based on color/feature histograms and our models based on the idea of complete or partial palette invariance.

Consider a collection of $T$ $I \times J$ signals $\mathbf{X} = \{X^t\}_{t=1}^T$, where each signal is defined on the domain $\{(i,j)|i \in [1..I], j \in [1...J]\}$, and the individual pixels (or other measurements) are indexed by $t$, $i$, and $j$, i.e., $\mathbf{X}^t = \{\mathbf{x}_{ij}^t\}_{(1,1)}^{(I,J)}$. The following discussion requires at least two different indices, but we use three to denote the structure in the data we processed in our experiments. In the case of images, we assume that $t$ denotes an image in the set we are analyzing, and $i, j$ are pixel indices. In the case of audio signals, we consider again $t$ to denote different utterances we are classifying, while $i, j$ denote the coordinates in the spectrogram of an utterance, i.e., frequency index and frame index, respectively.

Despite the redundancy, we temporarily associate a separate measurement palette $\mathcal{C}_{i,j}^t$ and an index $s_{ij}^t$ with each local measurement $\mathbf{x}_{ij}^t$. We will later make the signals share the index maps but not the palettes (Fig. 1). Each palette $\mathcal{C}$ is a table of $S$ color, feature, energy, or other measurement models, indexed by $s$. For example, $\mathcal{C}(s) = \boldsymbol{\mu}_s$ could be an $[r, g, b]^T$ vector for the $s$th color in the table, as customary. Then, the color of a pixel is $\mathbf{x}_{ij}^t = C(s_{ij}^t)$.

Furthermore, we can think of each measurement model $\mathcal{C}_{ij}^t(s)$ as the parameters of a distribution $p(\mathbf{x}|\mathcal{C}(s))$ over all possible measurements, $\mathbf{x}$. For example, $\mathcal{C}(s)$ could be defined as the mean $\boldsymbol{\mu}_s$ and the covariance matrix $\boldsymbol{\Phi}_s$ of a Gaussian distribution over the observation $\mathbf{x}_{ij}^t$,

$$p(\mathbf{x}_{ij}^t|\mathcal{C}_{ij}^t(s_{ij}^t)) = \mathcal{N}(\mathbf{x}_{ij}^t; \boldsymbol{\mu}_s, \boldsymbol{\Phi}_s), \qquad (1)$$

where $\mathbf{x}_{ij}^t$ could be a vector with the color coordinates in a suitable color space, or a vector of Gabor coefficients, or a vector of quantitative texture descriptors, vector of spatial and temporal derivatives, energy measurement in a speech spectrogram, cepstrum coefficient, or any other vector describing a signal at a particular location.

First, we show how some of the traditional image and speech representations map into our notation.

**Color palette and image compression.**

In many image formats, it is assumed that each image has its own color table, i.e.,

$$\begin{aligned}
\mathcal{C}_{11}^1 = \mathcal{C}_{12}^1 = ... = \mathcal{C}_{I,J-1}^1 = \mathcal{C}_{IJ}^1 &\quad = \quad \mathcal{C}^1 = \{\boldsymbol{\mu}_s^1\}_{s=1}^S \\
\mathcal{C}_{11}^2 = \mathcal{C}_{12}^2 = ... = \mathcal{C}_{I,J-1}^2 = \mathcal{C}_{IJ}^2 &\quad = \quad \mathcal{C}^2 = \{\boldsymbol{\mu}_s^2\}_{s=1}^S \\
&... \\
\mathcal{C}_{11}^T = \mathcal{C}_{12}^T = ... = \mathcal{C}_{I,J-1}^T = \mathcal{C}_{IJ}^T &\quad = \quad \mathcal{C}^T = \{\boldsymbol{\mu}_s^T\}_{s=1}^S.
\end{aligned}$$

This representation is useful when each image contains a relatively small number of colors, but sampled from a large portion of the color space. Then, a small number of colors, e.g., $S = 256$, are found that represent all the colors in the image most faithfully. Each entry in the palette is a 24-bit color, but in each location $i, j$ in the image, only the 8-bit index $s_{ij}$ is stored, yielding almost a three-fold compression, as the size of the palette is negligible in comparison with the size of the image. Usually, each image has a separate color palette, although the palettes can also be shared.

**Spatially-invariant color or feature distribution models** Lots of simple image understanding tools rely on color or feature histograms. Faced with the huge variability in the visual data, these algorithms typically assume that images or their portions are as similar as the distribution of colors present in them, and they ignore the spatial configuration of the colors. In our notation this idea can be summarize into an assumption that a collection of similar images shares the same color model for all pixels

$$\mathcal{C}^1_{11} = \mathcal{C}^1_{12} = ... = \mathcal{C}^T_{I,J-1} = \mathcal{C}^T_{IJ} \quad = \quad \mathcal{C}. \qquad (2)$$

This model is usually captured either by a mixture of Gaussians or a color/feature histogram.

**Speech HMMs** In speech models, signal is typically represented as a sequence of states, whose joint distribution is modeled by a Markov model. States, which we denote by $c_j$ are associated with frames (index $j$ in our notation) and describe Gaussian distributions over the frequency content (indexed by $i$ in our notation). We can replace the direct Gaussian observation model with the indexed model, so that $p(\mathbf{x}^t_{ij}|\mathcal{C}^t_{ij}(s^t_{ij})))$ decribes local measurments, as above, but $c^t_j$ influences the indices $s$ through $p(\{s^t_{ij}\}^I_{i=1}|c^t_j)$. Traditional HMMs (including mixture-tying models) can be seen as a special case where $p(\{s^t_{ij}\}^I_{i=1}|c^t_j)$ is deterministic.

## 3 Palette-invariant models

We can derive a new class of models that assume that indices $s$ *are* dependent on the coordinates $i, j$, but this information is shared across the collection of images. For example, if we assume that index $s_{ij}$ for each location in the image is shared across the entire collection

$$s^1_{ij} = s^2_{ij} = ... = s^t_{ij} = ... = s^{T-1}_{ij} = s^T_{ij} = s_{ij}, \quad (3)$$

and the palette $\mathcal{C}^t$ for each image is shared across all locations $i, j$,

$$\mathcal{C}^t_{11} = \mathcal{C}^t_{12} = ... = \mathcal{C}^t_{ij} = \mathcal{C}^t_{I,J-1} = \mathcal{C}^t_{IJ} = \mathcal{C}^t, \quad (4)$$

we obtain a basic palette-invariant model which assumes a fixed spatial arrangement configuration of the features, but the features themselves can arbitrarily change from one image to the next. For example, Fig. 2 shows an index map that describes a whole class of objects. The index map was learned from 50 examples of car images, using the algorithm we will describe shortly. In the same figure, we show the inferred palettes for 8 detected car images outside our training set. One useful property of the palette-invariant model is that it equates the images taken under different overall level of illumination, however, as shown in our example, the objects with different surface properties but similar spatial structure are also considered similar under this model. The basic palette-invariant model can be extended in several ways, but the most important concept that we would like to focus on in this paper is the introduction of the variability in the index map.

**Modeling uncertainty: probabilistic index maps (PIM) .** We can relax the hard assumption in (3) and allow the indices that model the same location in the image to vary, but follow the same distribution

$$p(s^1_{ij} = s) = p(s^2_{ij} = s) = ... = p(s^T_{ij} = s) = p_{ij}(s), \qquad (5)$$

where location-dependent distributions $p_{ij}$ describe the levels of variability in different locations of the image, and the overall distribution over the index maps $S = \{s^t_{ij}\}$ is

$$p(S) = \prod_{i,j,t} p_{ij}(s^t_{ij}) \qquad (6)$$

For example, if the image collection $\mathbf{X}$ are the frames from a video of a tree that moves slightly in the wind while the illumination conditions are varying considerable due to the cloud movement, then added level of variability in the index map $(p_{ij})$ helps capture the flutter of the leaves, still allowing generalization of the image under varying illuminations. This variability is separate from the intra-image appearance variability captured in the individual palettes, and tends to model intra-class structural variability instead.

Of course, probabilistic index map models $p(S)$ can be more complex, for example, they can encode extra dependencies among indices $s^t_{ij}$. We will later show this in a speech model.

**Free energy of a *probabilistic* index map (PIM)**

Here, we derive the inference (E step) and the parameter update rules (M step) for the model that uses variable index map described by (5) and sample-independent color maps (4). In this model, each observation has a separate index $s^t_{ij}$ but the prior $p_{ij}(s)$

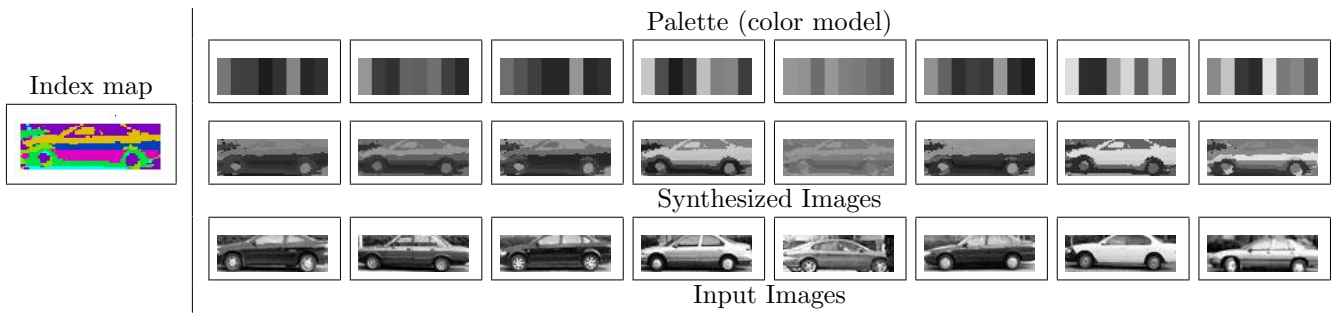

Figure 2: **A palette-invariant Generative model** In this model pixel values at every pixel in each input image (the third row) are determined by the acording one of possible distribution (an entry in the palette). Every image has its one palette (which can be approximated from the data)The decision which palette entry to chose is dictated by the global index map (left side)

for each location is shared among the observed images. The free energy is

$$F = \sum_{\mathbf{s}} q(\{s_{ij}^t\}) \log q(\{s_{ij}^t\}) - \qquad (7)$$
$$\sum_{\mathbf{s}} q(\{s_{ij}^t\}) \log \big[\prod_{i,j,t} p(\mathbf{x}_{ij}^t|s_{ij}^t, \mathcal{C}^t) \prod_{ij} p_{ij}(s_{ij}^t)\big].$$

The true posterior is factorized, so we can set $q(\{s_{ij}\}) = \prod_{i,j} q(s_{ij})$ with the bound being tight. Thus,

$$\begin{aligned}F &= \sum_{i,j,t}\sum_{s_{ij}^t} q(s_{ij}^t) \log q(s_{ij}^t) - \\ &- \sum_{i,j}\left(\sum_t q(s_{ij}^t)\right) \log p_{ij}(s_{ij}) - \\ &- \sum_{i,j,t}\sum_{s_{ij}^t} q(s_{ij}^t) \log p(\mathbf{x}_{ij}^t|s_{ij}^t, \mathcal{C}^t). \qquad (8)\end{aligned}$$

The resulting E step that optimizes the bound wrt q is

$$q(s_{ij}^t) \propto p_{ij}(s_{ij}^t) p(\mathbf{x}_{ij}^t|s_{ij}^t, \mathcal{C}^t). \qquad (9)$$

Assuming a Gaussian model in each entry of the palette $\mathcal{C}^t(s) = \{\boldsymbol{\mu}_s^t, \boldsymbol{\Phi}_s^t\}$, we obtain the following update rules for the M step:

$$p_{ij}(s_{ij} = s) = \frac{1}{T}\sum_t q(s_{ij}^t = s) \qquad (10)$$

$$\boldsymbol{\mu}_s^t = \frac{\sum_{ij} q(s_{ij}^t = s)\mathbf{x}_{ij}^t}{\sum_{ij} q(s_{ij}^t = s)} \qquad (11)$$

$$\boldsymbol{\Phi}_s^t = \frac{\sum_{ij} q(s_{ij}^t = s)[\mathbf{x}_{ij}^t - \boldsymbol{\mu}_s^t][\mathbf{x}_{ij}^t - \boldsymbol{\mu}_s^t]^T}{\sum_{ij} q(s_{ij}^t = s)}$$

We show these equation to emphasize the fact that the index maps are inferred based on the general tendency of the measurements to cluster together, but the actual cluster means and variances are estimated independently for different portions of the data, e.g., different images can have completely different palettes but the inferred index values tend to be the same.

## 4 Probabilistic index maps in complex graphical models

The factorized form of the posterior discussed in the previous section is actually exact when there are no additional hidden variables, but in general, factorization can be used as an approximation not only to make inference and learning more tractable for complex graphical models, but also to modularize the inference engine. In this section, we develop two examples of complex graphical models.

The first model captures image structure using probabilistic index map representation, but has an additional hidden variable denoting image transformation, as well as an image class. Thus, learning in this model allows for unsupervised clustering of images that is both transformation- and palette-invariant.

The second model captures speech signals in a way that is invariant to various types of signal noise that change the distribution over measurements in utterances.

### 4.1 Transformed mixtures of probabilistic index maps

Adding both the mixing variable $c$ and the transformation variable $\mathbf{T}$, we can construct a transformed mixture of PIMs (TMPIM), with the joint probability distribution for the $t$-th image

$$p(\mathbf{X}^t, S^t, c^t, \mathbf{T}^t) = p(\mathbf{X}|\mathbf{T}^t, S^t)p(S^t|c^t)p(T^t)p(c^t). \quad (12)$$

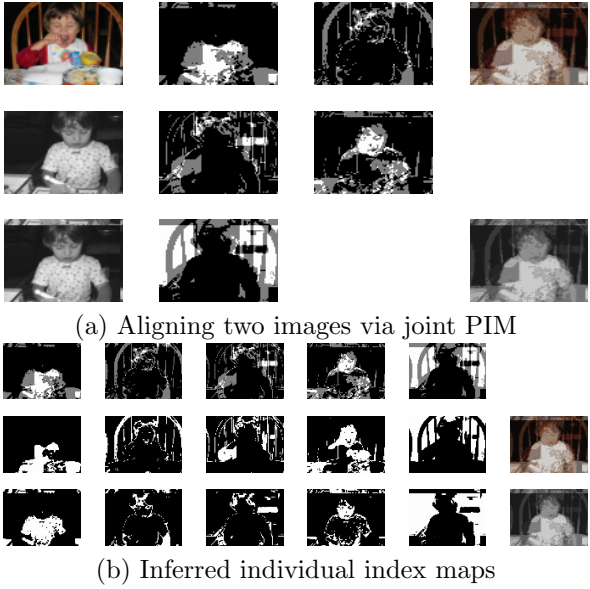

(a) Aligning two images via joint PIM

(b) Inferred individual index maps

Figure 3: Aligning two images with different colors and features. The first two images in the first column of (a) show two different images of a child taken on different days. One image is in color and the other one is black and white. The third image int he column shows the result of BW image alignment. The middle two columns show the probabilistic index map in terms of its components $p(s_{ij} = k)$ using a palette with only five entries. The last column shows the probabilistic index map in terms of the palettes inferred in two images. In (b), we contrast the probabilistic index map $p(s_{ij})$ (top row) which describes both images and the inferred index maps $q^t(s_{ij} = k)$ for individual images.

The observation distribution is defined as

$$\log p(\mathbf{X}|S,\mathbf{T}) = \\ -\tfrac{1}{2}\sum_{ij}\big[(\mathbf{x}_{ij}-\boldsymbol{\mu}_{s_{\mathbf{T}(ij)}})'\boldsymbol{\Phi}^{-1}_{s_{\mathbf{T}ij}}(\mathbf{x}_{ij}-\boldsymbol{\mu}_{s_{\mathbf{T}(ij)}}) \\ + \log|2\pi\boldsymbol{\Phi}_{s_{\mathbf{T}ij}}|\big], \quad (13)$$

where $\mathbf{T}(ij)$ are the coordinates into which $ij$ maps under $\mathbf{T}$. If not handled properly, this part of the generative model will be the main source of intractability, as maximizing it jointly over color distribution parameters ($\boldsymbol{\mu}_s, \boldsymbol{\Phi}_s$) and transformations $\mathbf{T}$ will be required. To transform the above into a more tractable computation, we rearrange the summation so that we first sum over all pixels that map to color $s = 1$, then as those that map to color $s = 2$, and so on:

$$\log p(\mathbf{X}|S,\mathbf{T}) = \\ -\tfrac{1}{2}\sum_{k=1}^{S}\sum_{i,j|s_{\mathbf{T}(ij)}=k}\big[(\mathbf{x}_{ij}-\boldsymbol{\mu}_k)'\boldsymbol{\Phi}^{-1}_k(\mathbf{x}_{ij}-\boldsymbol{\mu}_k) \\ +\log|2\pi\boldsymbol{\Phi}_k|\big], \quad (14)$$

Without the loss of generality, and for the sake of notational simplicity, we will focus on the case of a gray level (scalar) pixels [1], in which case we can write

$$\log p(\mathbf{X}|S,\mathbf{T}) = -\frac{1}{2}\sum_{k=1}^{S} d_k, \quad (15)$$

$$d_k = \mathbf{T}(S_k)'[\phi_k^{-1}(\mathbf{X}-\mu_k)^2 + \log 2\pi\phi_k]. \quad (16)$$

where we use $S_k$ to denote the binary image indicating for each pixel if it is assigned to palette entry $k$ or not, and $\mathbf{T}(S_k)$ is the transformed version of this binary image. These binary images and image $\mathbf{X}$ are represented as one-dimensional vectors of pixels (unwrapped images), so that the distance $d_k$ can be written as an inner product. Palette entry parameters $\mu_k, \phi_k$ are scalar, and the sum of a vector and a scalar is defined as adding the scalar to all elements of the vector, i.e., $\mathbf{X}+\mu = \mathbf{X}+\mu\mathbf{E}$, where $\mathbf{E}$ is the vector of ones. With this transformation of the observation likelihood and the variational approximation of the posterior,

$$q^t = q(c^t)q(S^t|c^t)q(\mathbf{T}^t|c^t), \quad (17)$$

$$q(S^t|c^t) = \prod_{ij} q(s^t_{ij}|c^t), \quad (18)$$

The cost of inference in a PIM-based transformed model is linear in the number of entries in the palette. Typically, a very small number of entries is sufficient (5-8). In addition, the treatment of the transformational variability in inference can be based on FFTs as in our previous work or usual multiresolution approaches used in vision. We feel that for the UAI audience we can omit the update rules for brevity. The iterative optimization consists of iterating (a) optimization of the color or feature palette ($\{\boldsymbol{\mu}_k,\boldsymbol{\phi}_k\}$) for each image; (b) inference of the variational posterior for each image (posterior distribution over the segmentation map, transformation that aligns the image with the current guess at PIM $p(S|c)$, and the posterior distribution over the class $c$ for each image); and (c) re-estimation of the class PIMs $p(s|c)$ and the prior $p(c)$. All of these steps are performed by minimizing the free energy of the model.

The resulting algorithm is illustrated on a mini two-image dataset in Fig. 3. In order to align a color image of a child with another gray-level image, we train a single-class TMPIM model which brings the two images into alignment with respect to the shared probabilistic index map. An example of unsupervised clustering images with TMPIM is given in the experimental section.

---

[1]When the covariance matrix $\boldsymbol{\Phi}_k$ is diagonal, the Mahalanobis distance breaks into a sum of distances between scalars, and if $\boldsymbol{\Phi}_k$ is not diagonal, it can be diagonalized by SVD, so both cases can be reduced to the case of scalar observations $x_{ij}$.

## 5 PIM in speech HMM models

As mentioned above, PIM can be used instead of a direct Gaussian observation models in HMMs for speech recognition. If we denote the hidden states by $c_j^t$, for the $t$-th utterance in a training set, then the PIM $p(S)$ can be modeled as

$$p(S) = \prod_j p(s_{ij}^t | c_j^t) \qquad (19)$$

The energy palettes in the spectrogram are shared across frames but are different in different frequency bands, i.e., $\mathcal{C}_{ij}^t = \mathcal{C}_i^t$, allowing the model to be invariant to various types of frequency-dependent noise.

The model is trained using a variational posterior that decouples joint distribution $q(c_1^t, .., c_J^t)$ over *all* states from the posterior distribution over the indices $\prod_{i,j} q(s_{ij}^t)$, leading to an efficient estimation that alternates between a forward backward algorithm on the state sequence and the palette and index inference.

## 6 Experimental results and conclusions

**Vision examples** The probabilistic index map is a universal image representation that can find its way into many existing computer vision algorithms. To illustrate its benefits, we applied the representation in two typical computer vision tasks: background subtraction and transformation-invariant image clustering.

In Fig. 4 we show that PIM representation allows for background subtraction based on a single frame, rather than on tracking incremental changes in a continuous video stream. An 8-index PIM model is learned by minimizing its free energy on the small collection of background images (Section 3). Then, for each new test image, the color palette is inferred and the pixel-wise free energy $F_{ij}^t$ is estimated. The foreground detection is then given in terms of the bumps in the energy profile, shown in the last column in (c-f). To better illustrate what is happening "under the hood," the middle column shows the expected background image $\mathbf{B}^t$ using the inferred color palette for each test image,

$$E[b_{ij}^t] = \sum_k q(s_{ij}^t = k) \boldsymbol{\mu}_k. \qquad (20)$$

Note, however, that the free energy also depends on the inferred variance for each palette entry.

It is interesting to compare the use of PIM representation with standard appearance models in more complex graphical models. Here, we use Frey and Jojic' transformed mixtures of Gaussians (TMG) for comparison, as this model captures variability in both appearance and transformation, and has been shown to be successful at unsupervised image clustering [2]. This model has been extended to capture dynamic properties of the scene as well as multiple layers of objects [3]. Sections 3 and 4 illustrate that our probabilistic index map representation can be used in similar situation as a simple pixel-wise Gaussian model, and so our new models should be equally easily extensible as TMG. However, as we show in Fig. 5, PIM representation leads to superior illumination invariance at a low extra computation cost. Using the 200 images from the dataset published with the TMG algorithm (Fig. 4a in [2]), the transformed mixture of probabilistic index maps was able to automatically cluster the data in (a) into two clusters representing two different people with an error rate of only 2.5%. In contrast regular mixture of Gaussians and TMG had much poorer error rates of 40.5% and 26%, respectively. All three techniques were applied in a completely unsupervised fashion.

In all image experiments we report in this paper, we only used color or gray-level intensity as image features. This makes it easier to separate the benefits of the probabilistic image map representation from the wise choice of local image features. To use other features, we can form an extended feature vector that concatenates the color information with other local measurements. As the model allows for learning the covariance structures $\boldsymbol{\phi}_k$ for various entries in the palette, (or even more complex probability distributions), the feature selection occurs naturally, and some indices may tend to model smooth areas of uniform color, while others will capture uniformity in texture features, despite the high variance in color. In our future research, we plan to apply this approach on unsupervised clustering of photographs.

**Speech recognition** We trained 11 digit models on the total of 22 noise-free utterances from the same speaker in the Aurora database. In order to properly compare to the basic HMM model, we first trained a standard HMM model with various number of states, until we found that it performs the best with 10 states, yielding a recognition error of 32.5% on the test set with 44 utterances, which had 4 different types of noise (non-white, i.e., the noise affected different parts of the spectrogram differently). Then, we trained a PIM-based HMM with 10 states as well on the same training data. The PIM palettes, which are reestimated for each training and test sample, had 7 entries in total. The recognition error rate on the test data was a substantially lower 18%.

In both vision and speech examples, the improvement over the baseline is not due to increased complexity of the PIM-based models. In the speech example, for instance, we tried HMMs with various numbers of the states and compared with the best one. In the image clustering example, if the number of clusters in the model is increased, the TMG model starts separating the two faces only after the complexity is substantially higher than the TMPIM model's complexity. More importantly, the TMG model with a large number of classes has no way of identifying which of the clusters model the same face.

## 7 Conclusions

We have presented a novel representation of real-valued natural signals in which the observed signal values are assumed to share a palette locally (e.g., in a single image), while the indexing structure is shared globally, e.g., across a collection of images. The distribution over the indices is assumed to encode the spatial and/or temporal signal structure for an object class, while the actual measurement palettes are treated as hidden variables that change freely from signal to signal. This is in stark contrast with the usual approaches to modeling real-valued signals. For instance, in previous work, we modeled the basic appearance change using mixtures of Gaussians or factor analysis and added additional variables to capture spatial transformations of the appearance. Others have derived similar methods, with reasonable success, e.g., [1]. In this paper, we propose to use density functions over real values only to capture local measurements, while the signal structure for a certain class is defined as a distribution over discrete variables. Our experiments indicate that this strategy leads to an increased invariance to non-structural changes, such as illumination change, non-white audio noise, and even complex changes such as re-painting various surfaces of the object. We plan to test our models with different types of measurements, e.g, texture features rather than colors in images, as well as to use different probability models for the joint distribution over indices $p(S)$.

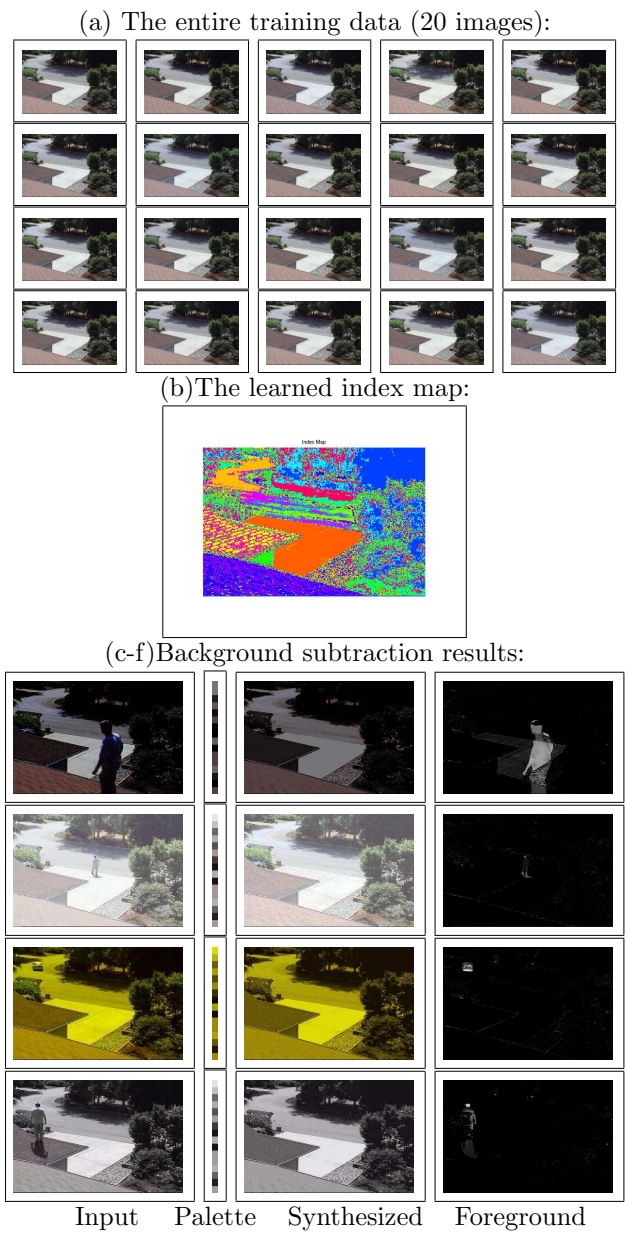

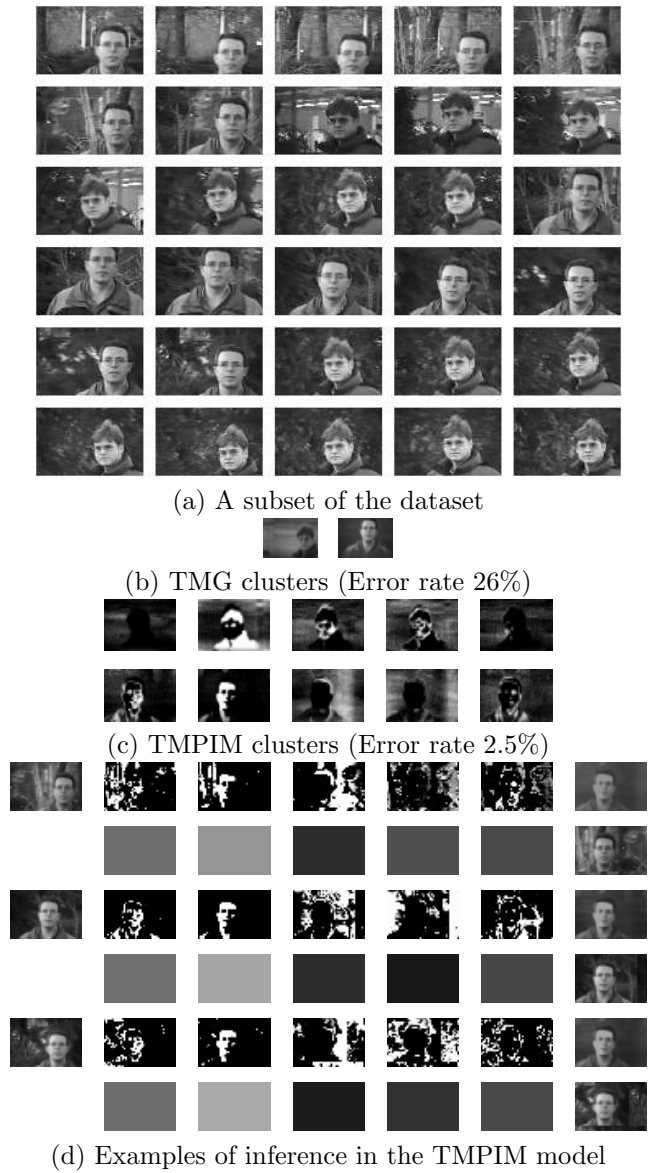

Figure 4: Illumination-invariant background subtraction. The background model is trained using only the 20 images shown in (a). The learned index map is shown in (b). Rows (c)-(d) show the images with drastic illumination changes, the recomputed background to match the the new conditions and the result of the background subtraction. The situations PIM model can handle include low illumination (c), image saturation (d), color channel malfunction (e), or even a switch to a different set of measurements, such as IR, or as in (f), gray-level images. Note that in all cases the recovery from the illumination change is instantaneous, and that the color training data had no examples remotely similar in intensities to the test examples.

Figure 5: Unsupervised clustering using transformed mixtures of probabilistic index maps (TMPIM). TMPIM clusters are represented by two distributions $p(S|c)$, shown as probability maps for index k=1,...,5. TMPIM, with its clustering accuracy of 97.5%, compares favorably to the standard mixture of Gaussians model that had a clustering accuracy of only 59.5% and the TMG technique [2] with accuracy of 74%. In (b) we show inferred variational posterior $q(S|c)$, the palette means, the synthesized image and the aligned input for three images.